\documentclass[sigconf]{acmart}
\usepackage{url,hyperref,lineno,microtype,subcaption}
\usepackage{xcolor,graphicx,times,siunitx,multirow}
\AtBeginDocument{%
  \providecommand\BibTeX{{%
    \normalfont B\kern-0.5em{\scshape i\kern-0.25em b}\kern-0.8em\TeX}}}

\setcopyright{acmcopyright}
\copyrightyear{2021}
\acmYear{2021}
\acmDOI{10.1145/1122445.1122456}

\acmConference[Bangalore '21]{Bangalore '21: ACM India Joint International Conference on Data Science \& Management of Data}{January 02--04, 2021}{Bangalore, India}
\acmBooktitle{Bangalore '21: ACM India Joint International Conference on Data Science \& Management of Data, January 02--04, 2021, Bangalore, India}
\acmPrice{15.00}
\acmISBN{978-1-4503-XXXX-X/18/06}

\begin{document}

\title{Beyond Textual Data: Predicting Drug-Drug Interactions\\ from Molecular Structure Images using\\ Siamese Neural Networks}

\author{Devendra Singh Dhami}
\email{devendra.dhami@utdallas.edu}
\affiliation{%
  \institution{The University of Texas at Dallas}}

\author{Siwen Yan}
\affiliation{%
  \institution{The University of Texas at Dallas}}
\email{siwen.yan@utdallas.edu}

\author{Gautam Kunapuli}
\email{gautam.kunapuli@verisk.com}
\affiliation{%
 \institution{Verisk Analytics, Inc.}}

\author{David Page}
\email{david.page@duke.edu}
\affiliation{%
 \institution{Duke University}}

\author{Sriraam Natarajan}
\email{sriraam.natarajan@utdallas.edu}
\affiliation{%
  \institution{The University of Texas at Dallas}}

\renewcommand{\shortauthors}{Dhami et al.}

\begin{abstract}
Predicting and discovering drug-drug interactions (DDIs) is an important problem and has been studied extensively both from medical and machine learning point of view. Almost all of the machine learning approaches have focused on text data or textual representation of the structural data of drugs. We present the first work that uses drug structure images as the input and utilizes a Siamese convolutional network architecture to predict DDIs.
\end{abstract}



\maketitle

\section{Introduction}
Adverse drug events (ADEs) are ``injuries resulting from medical intervention related to a drug'' \citep{NebekerBS04}, and are distinct from medication errors (inappropriate prescription, dispensing, usage etc.) as they are caused by drugs at normal dosages. According to the National Center for Health Statistics \citep{CDC14}, 48.9\% of Americans took at least one prescription drug in the last 30 days, 23.1\% took at least three, and 11.9\% took at least five. These numbers rise sharply to 90.6\%, 66.8\% and 40.7\% respectively, among older adults (65 years or older). This means that the potential for ADEs is very high in a variety of health care settings including inpatient, outpatient and long-term care settings. For example, in inpatient settings, ADEs can account for as many as one-third of hospital-related complications, affect up to 2 million hospital stays annually, and prolong hospital stays by 2--5 days \citep{HHS10}.

\begin{figure*}[!t]
    \centering
    \includegraphics[width=\textwidth]{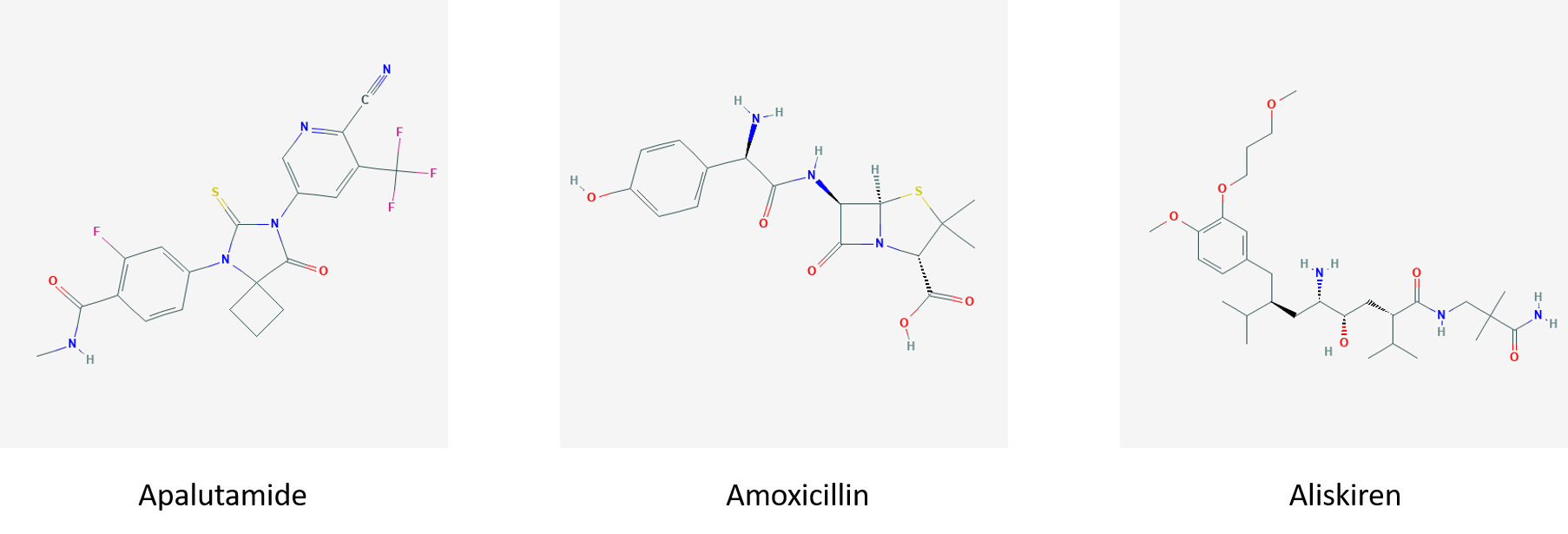}
    \caption{Some example molecular images of different drugs extracted from the PubChem database.}
    \label{fig:examples}
\end{figure*}

The economic impact of these issues is as widespread as the various healthcare settings and can be staggering. Estimates suggest that ADEs contributed to \$3.6 billion in excess healthcare costs in the US alone \citep{IMC06}. Unsurprisingly, older adults are at the highest risk of being affected by an ADE, and are seven times more likely than younger persons to require hospital admission \citep{BudnitzEtAl06}. In the US, as a large number of older adults are Medicare beneficiaries, this economic impact is borne by an already overburdened Medicare system and ultimately passed on to taxpayers and society at large. Beyond older adults, there are several other patient populations that are also vulnerable to ADEs including children, those with lower socio-economic means, those with limited access to healthcare services, and certain minorities. 

Recent research has identified, somewhat surprisingly, that many of these ADEs can be attributed to {\bf very common medications} \citep{BudnitzEtAl11} and {\bf many of them are preventable} \citep{GurwitzEtAl03} or {\bf ameliorable} \citep{ForsterEtAl05}. This issue motivates our long-term goal of developing accessible and robust means of identifying ADEs in a disease/drug-agnostic manner and across a variety of healthcare settings. Here, we focus on the problem of {\bf drug-drug interactions} (DDIs), which are a type of ADE. An ADE is characterized as a DDI when multiple medications are co-administered and cause an adverse effect on the patient. DDIs, often caused by inadequate understanding of various drug-drug contraindications, are a major cause of hospital admissions, rehospitalizations, emergency room visits, and even death \citep{becker2007hospitalisations}.

Predicting and discovering drug-drug interactions (DDIs) is an important problem and has been studied extensively both from medical and machine learning point of view. Identifying DDIs is an important task during drug design and testing, and regulatory agencies such as the U. S. Food and Drug Administration require large controlled clinical trials before approval. Beyond their expense and time-consuming nature, it is impossible to {\bf discover} all possible interactions during such clinical trials. This necessitates the need for computational methods for DDI prediction. A substantial amount of work in DDI focuses on text-mining \citep{liu2013azdrugminer,chee2011predicting} to extract DDIs from large text corpora; however, this type of information extraction does not discover new interactions, and only serves to extract {\em in vivo} or {\em in vitro} discoveries from publications. 

Our goal is to {\bf discover} DDIs in large drug databases by exploiting various properties of the drugs and identifying patters in drug interaction behaviors. Almost all of the machine learning approaches have focused on text data or textual representation of the structural data of drugs \citep{gurulingappa2012extraction,asada2018enhancing,purkayastha2019drug}. Recent approaches consider phenotypic, therapeutic, structural, genomic and reactive properties of drugs \citep{cheng2014machine} or their combinations \citep{dhami2018drug} to characterize drug interactivity. We take a fresh and completely new perspective on DDI prediction through the lens of molecular images, a few examples shown in figure \ref{fig:examples}, via deep learning. Our work is novel in the following significant ways:
\begin{itemize}
    \item we formulate DDI discovery as a {\bf link prediction problem};
    \item we aim to perform DDI discovery directly on {\bf molecular structure images} of the drugs directly, rather than on lossy, string-based representations such as SMILES strings and molecular fingerprints; and
    \item we utilize a deep learning technique, specifically {\bf Siamese networks} \citep{chopra2005learning} in a contrastive manner to build a {\bf DDI discovery engine} that can be integrated into a drug database seamlessly.
\end{itemize}

\section{Related Work}

\subsection{Drug-Drug Interactions}
The social and economic impacts of drug-drug interactions have also been well studied and understood. The effect of DDI on medication management and social care is studied in \citep{arnold2018impact} and with its economic impact shown in \citep{shad2001economic}. The impact of DDIs in the elderly patients in 6 Europen countries was documented in \citep{bjorkman2002drug} and in a similar vein the study by Becker et al. \citep{becker2007hospitalisations} identifies that the elderly have an increased risk factor $\approx$ 9 times over the general population with the clinical significance of DDIs studied in \citep{roberts1996quantifying}. Identification of DDIs can be done by either clinical trials or {\it in vitro} and {\it in vivo} experiments but these approaches are highly labor-intensive, costly and time-consuming. Thus, a system that can mitigate these factors is highly desirable.

Drug-Drug interactions have been studied extensively both from medical and machine learning point of view. From a medical standpoint \citep{lau2003atorvastatin}, \citep{hirano2006drug} and \citep{wang2000human} showed the effect of important individual drugs and enzymes such as subtrates on various drug-drug interactions. The problem of DDI discovery/prediction is a pairwise classification task and thus kernel-based methods \citep{Shawe-TaylorCristianini04} are a natural fit since kernels are naturally suited to representing pairwise similarities. Most similarity-based methods for DDI discovery/prediction have used biomedical research literature as the underlying data source and construct NLP-based kernels from these medical documents \citep{segura2011using,chowdhury2013fbk}. Some work has also been done on learning kernels from different types of data such as molecular and structural properties of the drugs and then using these multiple kernels to predict DDIs \citep{cheng2014machine,dhami2018drug}.\\ 

\subsection{Siamese Neural Networks}
Siamese networks have been applied in one shot image recognition ~\citep{koch2015siamese}, signature verification ~\citep{bromley1994signature}, 
object tracking ~\citep{bertinetto2016fully} and human re-identification ~\citep{varior2016gated,chung2017two}. Siamese networks have also been used in the health care domain in medical question retrieval \citep{wang2019medical} and Alzheimer disease diagnosis \citep{aderghal2017classification}. Siamese networks have also been used for the tasks of drug-drug interactions in the form of a Siamese graph convolutional network \citep{chen2019drug,jeon2019resimnet}.

\section{Siamese Convolutional Network for Drug-Drug Interactions}
A discriminative approach for learning a similarity metric using a Siamese architecture was introduced in \citep{chopra2005learning} which maps the input (pair of images in our case) into a target space such that the distance between the mappings is minimized in the target space for similar pair of examples and maximized in case of dissimilar examples.
\begin{figure*}[!t]
    \centering
    \includegraphics[width=\textwidth,height=9cm]{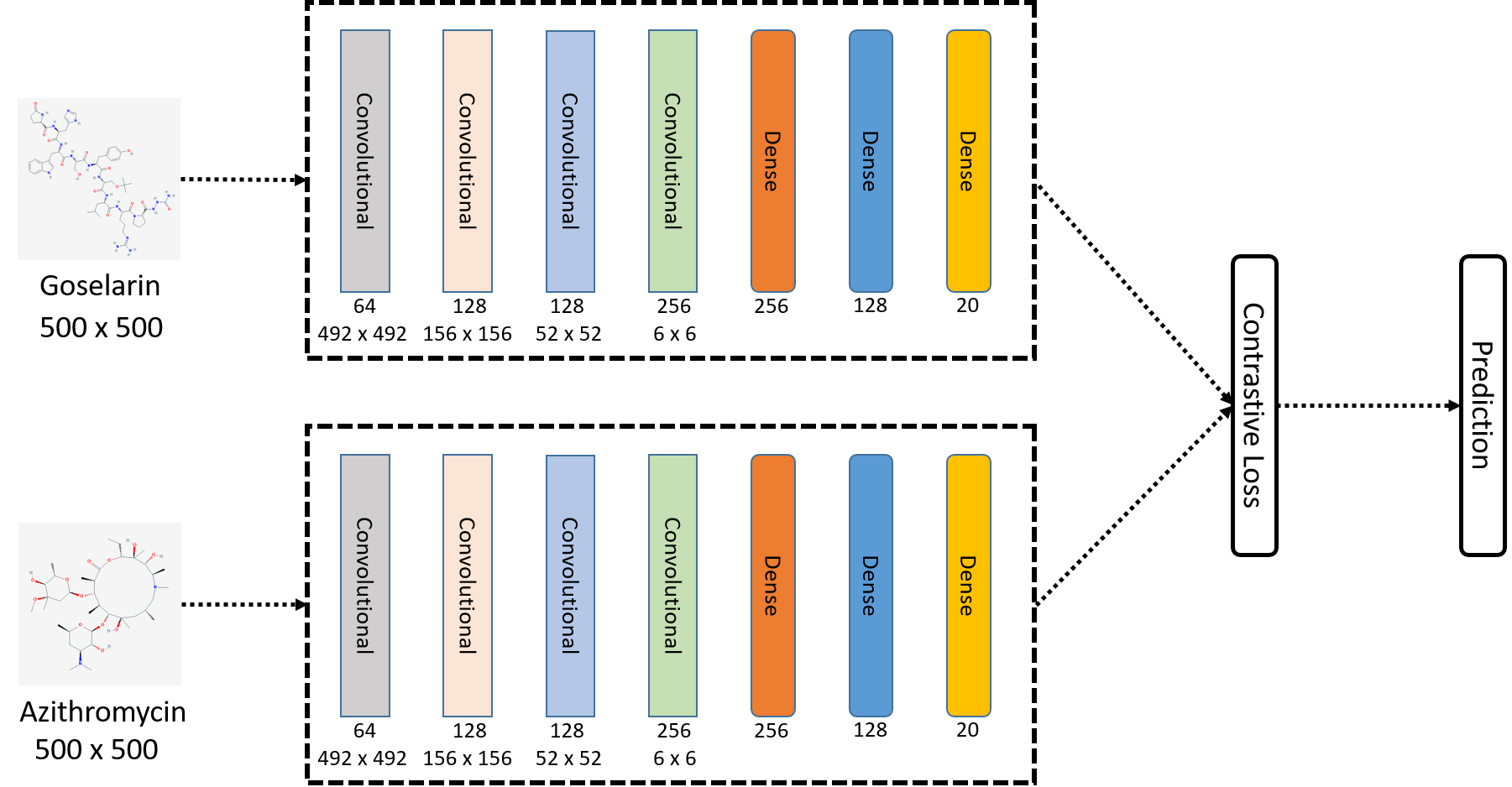}
    \caption{An overview of our model for predicting drug-drug interactions}
    \label{fig:overview}
\end{figure*}
We adapt the Siamese architecture for the task of link prediction where the link is whether two drugs interact or not. Since the Siamese architecture results in a measure of similarity between the pair of given inputs it can be thresholded in order to obtain a classification. We use contrastive loss \citep{hadsell2006dimensionality}, based on a distance metric (Euclidean distance in our case), to learn a parameterized function $F$ to obtain the mapping from the input space to the target space whose minimization can result in pushing the semantically similar examples together. 

An important property of the loss function is that it calculated on a pair of examples. The loss function is formulated as as follows: Let $X_1$ and $X_2$ are a pair of drug images and $Y$ is the label assigned to each of the pairs. The label $Y=0$ if the pair of drug images do not interact and $Y=1$ if the pair of drug images interact. Also, let $D$ be the Euclidean distance between the vector of the image pairs after being processed by the underlying Siamese network and $P$ are the parameters of the function F. The contrastive loss function can then be given as
\begin{equation}
    \mathcal{L}(P,X_1,X_2,Y)=\frac{(1-Y)}{2}{D_P}^2+\frac{Y}{2}\{\max(0, m-{D_P})\}^2
\label{eq}
\end{equation}
where $D_P={\|F_P(X_1) - F_P(X_2)\|_2^2}$ is the Eucledian distance between the obtained outputs after the input pairs are processed by the sub-networks. Also \textit{m} is a margin such that \textit{m} $\geq$ 0 that signifies that dissimilar pairs beyond this margin will not contribute to the loss.

Figure \ref{fig:overview} shows our complete architecture. It consists of two identical sub-networks i.e. networks having same configuration with the same parameters and weights. Each sub-network takes a gray-scale image of size 500 $\times$ 500 $\times$ 1 as input (we initially have color images that we convert to gray-scale before feeding to sub-networks as input) and consists of 4 convolutional layers with number of filters as 64, 128, 128 and 256 respectively. The kernel size for each convolutional layer is (9 $\times$ 9) and the activation function is \textit{relu}. The \textit{relu} is a non-linear activation function is given as $f(x)=max(0,x)$. Each convolutional layer is followed by a max-pooling layer with pool size of (3 $\times$ 3) and a batch normalization layer. After the convolutional layers, the sub-network has 3 fully connected layers with 256, 128 and 20 neurons respectively. Thus after an image pair is processed by the Siamese sub-networks two vectors of dimension 20 $\times$ 1 are obtained. Contrastive loss is then applied to the obtained pair of vectors to obtain a distance between the input pair which can then be thresholded to obtain a prediction. 

\begin{figure}[h!]
    \begin{center}
    \includegraphics[width=\columnwidth]{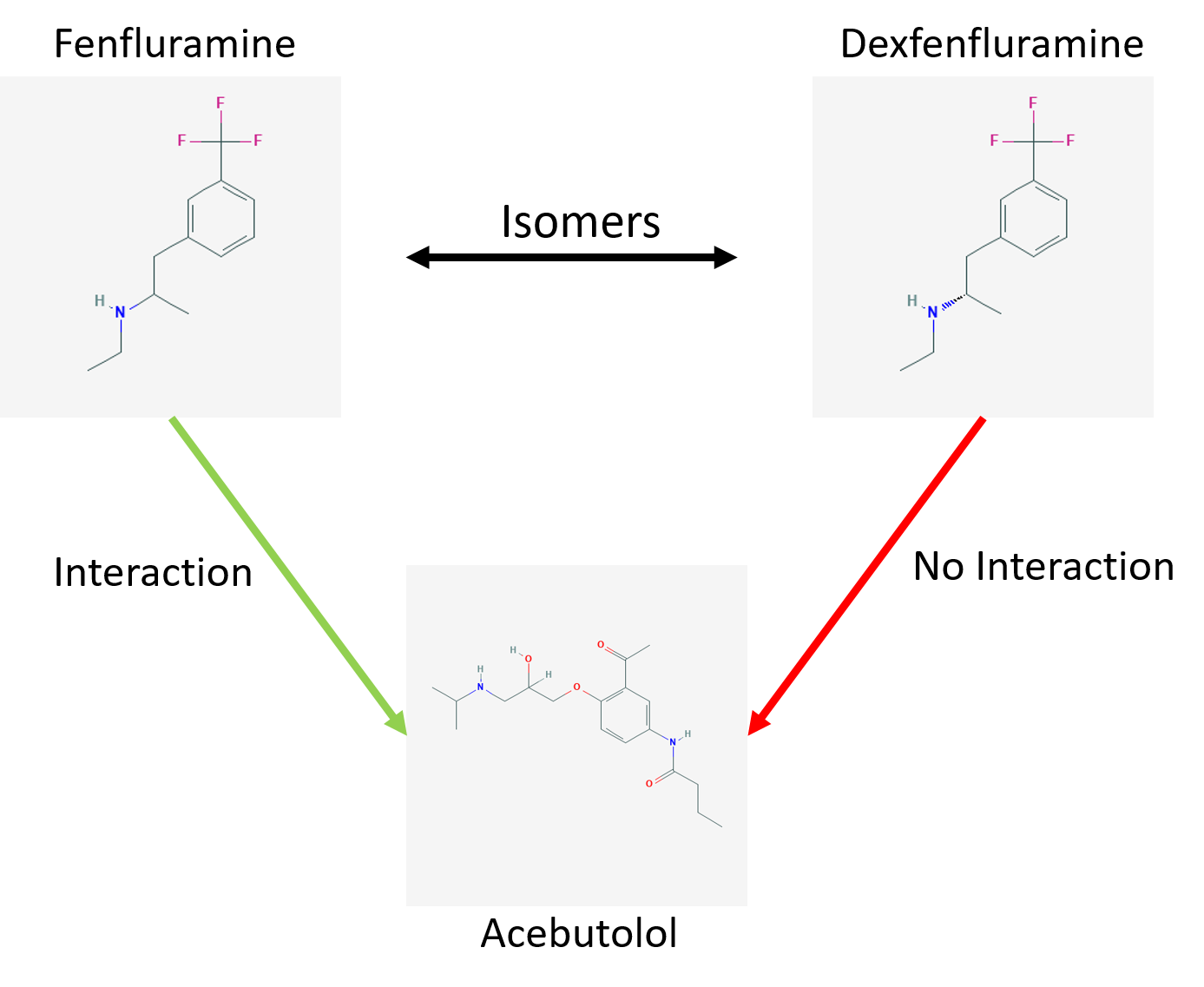}
    \end{center}
    \caption{An example of how two isomers interact differently with a single drug.}
    \label{fig:isomer}
\end{figure}

\begin{figure*}[h!]
    \begin{center}
    \includegraphics[width=\textwidth]{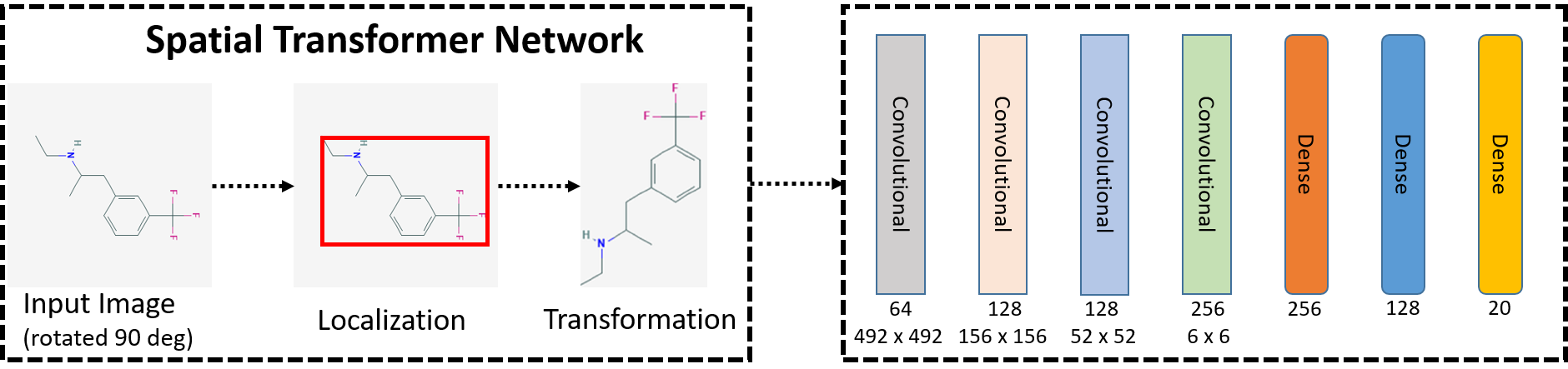}
    \end{center}
    \caption{Using spatial transformer network as a pre-processing step to mitigate rotational variance. Note that this process is done for both the input images.}
    \label{fig:stn}
\end{figure*}

We use the technique of precision recall curve (PR-curve) to identify the best threshold $=$ 0.65. Note that the convolutions in the convolutional sub-network provide translational in-variance property but rotational in-variance is also important in our problem domain. This is because isomers (one of the chiral forms) of drugs are expected to react differently when interacting with a certain drug \citep{nguyen2006chiral,chhabra2013review}. For example, Fenfluramine and Dexfenfluramine are isomers of each other and where Fenfluramine interacts with Acebutolol but Dexfenfluramine does not (Figure \ref{fig:isomer}). Another example is that the L-isomer of methorphan, Levomethorphan, is an opioid analgesic, while the D-isomer, Dextromethorphan, is a dissociative cough suppressant\footnote{https://en.wikipedia.org/wiki/Enantiopure\_drug}.  To overcome this problem and introduce rotational invariance into our framework, we make use of spatial transformer networks \citep{jaderberg2015spatial} that we discuss next.

\subsection{Spatial Transformer Networks}
Spatial Transformer Network (STN) is a visual attention mechanism that can handle the scaling and rotation of the input images to the underlying convolutional network thereby leading to a better performance by reducing the effect of the rotation variance, which is a hard problem for convolutional neural networks \citep{cohen2014transformation}. It consists of three basic building blocks: a localisation network, a grid generator and a sampler which can be used as a pre-processing step before feeding the input image pair into our underlying Siamese architecture as shown in Figure \ref{fig:stn}. The whole network is differentiable, which means that it can be plugged directly into an existing model. The localisation network is used to regress the transformation parameters $\theta$, which controls the rotation, translation, zooming in and zooming out of the input images. 

The localisation network takes the input image, say $X_1$ in our case, and generates $\theta = {f}_{loc}(X_1)$ that can then be used to calculate the target image $\hat{X_1}$ . There is no specific requirement for the localisation network except it should be able to generate regression value for $\theta$. Our localization network is a convolutional neural network consisting of 2 pooling layers, 2 convolutional layers and 2 dense layers. The transformation parameters $\theta$ is the mapping between source image coordinators $\left({x}^{X_1}_{i},{y}^{X_1}_{i}\right)$ and target image coordinators $\left({x}^{\hat{X_1}}_{i},{y}^{\hat{X_1}}_{i},1\right)$ as shown by the equation \ref{stn_theta}. Note that transformation function is not learned explicitly rather is learned automatically by the network. 

\begin{equation}
    \begin{pmatrix}
     {x}^{X_1}_{i}\\ 
     {y}^{X_1}_{i}
    \end{pmatrix}=
    \begin{bmatrix}
     {\theta}_{11} & {\theta}_{12} & {\theta}_{13}\\ 
     {\theta}_{21} & {\theta}_{22} & {\theta}_{23}
    \end{bmatrix}
    \begin{pmatrix}
     {x}^{\hat{X_1}}_{i}\\ 
     {y}^{\hat{X_1}}_{i}\\ 
     1
    \end{pmatrix}
\label{stn_theta}
\end{equation}

Hence, the localization and transformation as shown in Figure 4 are done in a single step. For the sampling kernel, we used the standard bilinear interpolation as described in \citep{jaderberg2015spatial}, since gradients can be defined with respect to the source image coordinates for bilinear interpolation.

\section{Experiments}
We aim to answer the following questions:
\begin{enumerate}
    \item[] \textbf{Q1:} Are Siamese networks effective in link prediction task of DDI?
    \item[] \textbf{Q2:} What is the effect of number of epochs on the predictive performance of the Siamese architecture?
    \item[] \textbf{Q3:} Does our architecture handle the problem of rotational variance?
    \item[] \textbf{Q4:} Are molecular structure images informative enough to predict DDIs and can be used instead of lossy string representations?
    \item[] \textbf{Q5:} How does our method compare with state-of-the-art statistical relational models?
    \item[] \textbf{Q6:} How does the choice of distance function for contrastive loss effect the prediction performance?
    \item[] \textbf{Q7:} How does the choice of optimization function effect the prediction performance?
\end{enumerate}

\subsection{Data set}
Our data set consists of images of 373 drugs of size 500 $\times$ 500 $\times$ 3 downloaded from the PubChem database \footnote{https://pubchem.ncbi.nlm.nih.gov/} and converted to a grayimage format to yield images of size 500 $\times$ 500 $\times$ 1. From these images we create a total of 67,360 drug interaction pairs excluding the reciprocal pairs (Since drug-drug interaction is reciprocal in nature i.e. if drug $d_1$ interacts with drug $d_2$ then $d_2$ interacts with $d_1$ and vice versa, we need to remove such pairs from our data). From the 67,630 drug pairs we obtain a data set of 19936 drug pairs that interact with each other ($Y$ = 1) and 47424 drug pairs that do not interact with each other ($Y$ = 0). The images are normalized by the maximum pixel value (i.e. 255) before passing to the network. The data set and the code is available at \url{https://rb.gy/koax5u}.
\begin{figure}[h!]
    \centering
    \includegraphics[width=\columnwidth]{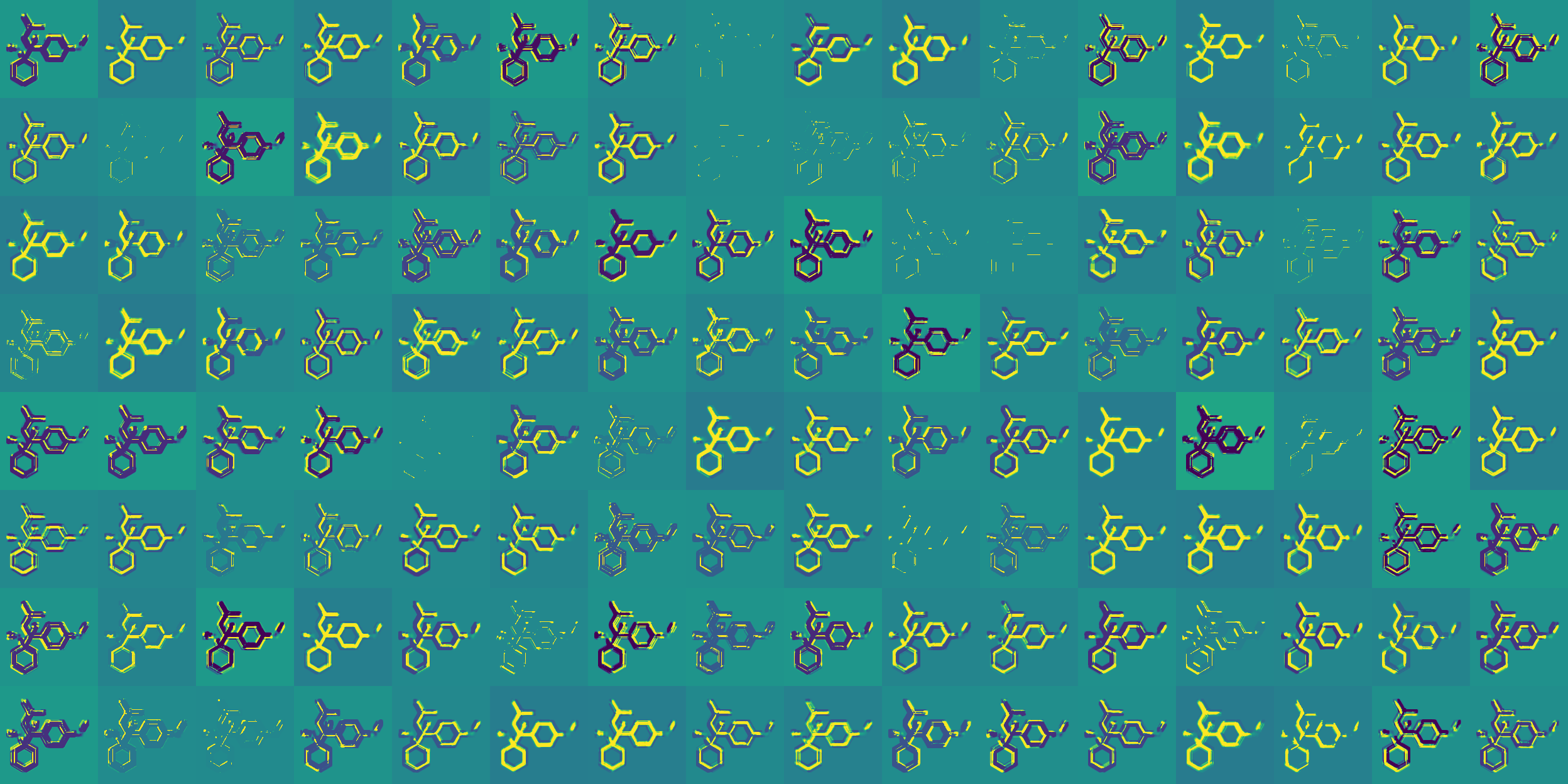}
    \caption{An example of abstract features learned by a convolutional layer for Venlafaxine.}
    \label{fig:venla_conv}
\end{figure}
\subsection{Baselines}
We consider 5 baselines using different data modalities to compare the results from our Siamese architecture, namely,
\begin{enumerate}
    \item \textit{\textbf{Image data:}}\\ \textbf{1. Structural Similarity Index (SSIM):} is used for measuring perceptual similarity between images \citep{wang2004image} and given 2 images $X_1$ and $X_2$ is calculated as,
    \begin{equation}
        SSIM(X_1,X_2)=\frac{(2 \mu_{X_1}  \mu_{X_2} + C_1) \times (2 \sigma_{X_1 X_2} + C_2)}{(\mu_{X_1}^2 + \mu_{X_2}^2 +C_1) \times (\sigma_{X_1}^2 + \sigma_{X_2}^2 +C_2)}
    \end{equation}
    where $\mu_{X_1}$ and $\mu_{X_2}$ is the average of the images $X_1$ and $X_2$ respectively, $\sigma_{X_1}$ and $\sigma_{X_2}$ is the variance of the images $X_1$ and $X_2$ respectively, $\sigma_{X_1 X_2}$ is the covariance of the two input images. The constants $C_1$ and $C_2$ are added to the SSIM to avoid instability and are the product of a small constant ($\ll$ 1) with the dynamic range of pixel values in the given images. The SSIM measure can also be written as the product of three types of comparisons between input images, namely, luminance, contrast and structure. To obtain the predictions, the SSIM needs to be thresholded and in the experiments, the threshold is set as the mean SSIM values of all pairs.\\
    \textbf{2. Autoencoders:} are neural networks that consists of 2 main components: an encoder and a decoder \citep{kramer1991nonlinear}. The encoder extracts features from the input images and decoder restores the original images from the extracted features. In general, the performance of autoencoders are evaluated by pixel-wise comparison between input images and output images. In order to compare the similarity between two images, the similarity between extracted features of the two images can be compared. This approach should be able to find images which contain objects with similar color and shape.
     
     For the encoder, we have three convolutional layers with filter sizes 16, 32 and 64 followed by a max pooling layer which is in turn followed by two convolutional layers with filter sizes 128, 64 and another max pooling layer. The final three convolutional layers consists of filters of sizes 32, 16 and 8. For the decoder, we have two convolutional layers with filter sizes 16 and 32 followed by a single up-sampling layer which is in turn followed by two convolutional layers with filter sizes 64 and 128 again followed by a single up-sampling layer. The final four convolutional layers consists of filters of sizes 64, 32, 16 and 1. The size of all kernels is 3$\times$3. The size of max pooling is 2$\times$2 and up sampling size is also 2$\times$2. The activation of all convolutional layers is relu, except the last layer of both encoder and decoder is a sigmoid, for the ease of comparison.
     
    First, the autoencoder model is trained using the training images, as is the normal training process of an autoencoder model. The number of epochs is 10 and the loss function is binary cross-entropy. Then features are extracted using the encoder on the testing images. To find images with similar extracted features, a couple of criterion were used, namely,  binary cross-entropy and cosine proximity. The threshold to decide whether the two images is similar or not was set as the mean of all values calculated for all pairs of testing image.
    \item \textit{\textbf{String data:}}\\ 1. \textbf{CASTER} \citep{huang2020caster} uses the drug molecular structure in a text format of \textbf{S}implified \textbf{M}olecular \textbf{I}nput \textbf{L}ine \textbf{E}ntry \textbf{S}ystem (\textbf{SMILES}) \citep{weininger1988smiles} strings representation to predict drug-drug interactions and ouperforms several deep learning methods such as DeepDDI \citep{ryu2018deep} and molVAE \citep{gomez2018automatic}. CASTER identifies the frequent substrings present in the SMILES strings presented during the training phase using a sequential pattern mining algorithm which are then converted to a emdedded representation using an encoder module to obtain a set of latent feature vectors. These features are then converted into linear coefficients which are then passed through a decoder and a predictor to obtain the DDI predictions. We obtain the SMILES strings of all the drugs in our data set from PubChem and DrugBank \footnote{https://www.drugbank.ca/} and use the source code \footnote{https://github.com/kexinhuang12345/CASTER} provided by the authors along with provided default hyper parameter settings.
    \item \textit{\textbf{Relational Data:}}\\ \textbf{1. RDN-Boost} \cite{natarajan2012gradient} extends the functional gradient boosting framework \cite{friedman2001greedy} to the relation setting by boosting relational dependency networks (RDNs) \cite{neville2007relational} with the aim to overcome the assumption of a propositional representation of the data as in standard functional gradient boosting. The objective function used in is the log-likelihood and probability of an example is represented as a sigmoid over the learned relational regression trees (RRT) \cite{blockeel1998top} which uses the relational features as input. The basic idea is to take an initial model (RRT) and use the obtained predictions to compute gradient(s) or residues. A new regression function i.e. a new RRT is then learnt to fit the residues and the model is updated. At the end, a combination (the sum) of all the obtained regression function gives the final model.\\
    \textbf{2. MLN-Boost} \cite{khot2011learning} boosts the undirected Markov logic networks (MLNs) \cite{richardson2006markov} instead of the directed relational dependency networks in case of RDN-Boost. In MLN-Boost the structure and parameters of the MLN are learned simultaneously by converting the problem of learning MLNs to a series of relational functional approximation problems similar to the RDN-Boost setting, with the only difference being that the number of groundings for each learned clause are counted in case of MLN-Boost whereas RDN-Boost uses existential semantics.
\end{enumerate}
We convert the data obtained from DrugBank to the relational format with number of relations = 14 and the total number of facts = 5366. For both RDN-Boost and MLN-Boost we set the number of relational regression trees to be learned as 10.
\subsection{Results}
We optimize our Siamese network using the Adam as the optimization algorithm \citep{kingma2014adam} with a learning rate of $5 \times 10^{-5}$ (we also train the network using several other optimization algorithms as defined later). The best learning rate was obtained using line search. We set the value of the margin $m$ in contrastive loss equal to 1. As mentioned before, we keep the threshold value as 0.65, obtained using AUC-PR curve, to obtain the predictions after obtaining a distance between pair of drug images using the Siamese convolutional network. We divide our data set into 44457 training (66\% of the data) and 22903 testing examples. Example features learned by the second convolutional layer in our network for the drug Venlafaxine is shown in figure \ref{fig:venla_conv}. When pre-processing the data using a STN, we rotate the data set images by \ang{90} and pass it through the STN before passing it through our Siamese network. Another important thing to note here is that in our problem formulation recall is the most important factor that should be considered. The simple reason is that we do not want to miss any interaction i.e. a false negative results in much more serious consequences (fatalities in patients) than false positives (monetary losses such as new clinical trials) \citep{dhami2018drug} \textbf{although a recall gain should not come at the cost of loss in precision} since that can be obtained simply by classifying every test example as a positive example.

\begin{figure}[h!]
    \centering
    \includegraphics[width=\columnwidth]{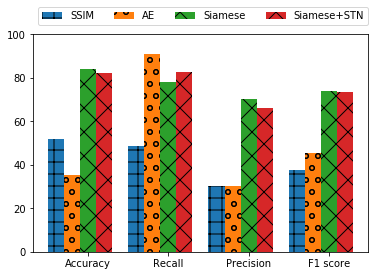}
    \caption{Results for DDI prediction using images. Although the recall of auto-encoders (AE) is higher than the Siamese network, low precision shows a high rate of false positives.}
    \label{fig:image_base}
\end{figure}


Figure \ref{fig:image_base} shows the results of using our Siamese network architecture with and without rotational invariance (STN) compared with baselines. The Siamese network with and without STN (results here reported for 50 epochs for both cases) outperforms the baselines thereby answering \textbf{Q1} affirmatively. Siamese networks are clearly effective and significantly better for the DDI task of link prediction. Note that although the recall of auto-encoders is higher than the Siamese network, the very low precision shows a high rate of false positives and thus its performance cannot be judged as being better than the proposed model.

\begin{figure}
\centering
\subcaptionbox{Performance of Siamese network across epochs\label{fig:siam_epoch}}
{\includegraphics[width=\columnwidth]{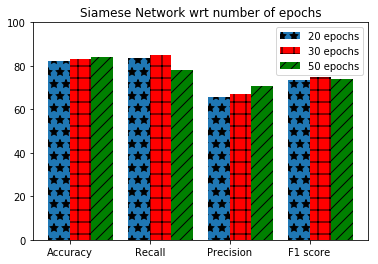}}
\subcaptionbox{Performance of Siamese network with STN across epochs\label{fig:stn_epoch}}
{\includegraphics[width=\columnwidth]{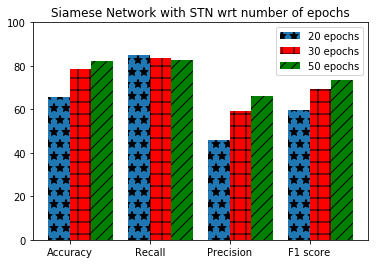}}
\caption{Variation of DDI prediction of the proposed networks wrt epochs.}\label{fig:epochs}
\end{figure}

Figure \ref{fig:epochs} shows the variation of performance of Siamese network (figure \ref{fig:siam_epoch}) and Siamese network with STN (figure \ref{fig:stn_epoch}) with respect to the number of iterations. The results for Siamese networks without STN do not show any significant change wrt the increasing epochs across metrics whereas in case of Siamese networks with STN, the results show a steady increase with increasing iterations across majority of metrics. The recall decreases with increasing number of epochs in both cases i.e Siamese networks with and without STN but the decrease is more stark in case of the network without STN whereas the drop is not significant in the other case with STN. This answers \textbf{Q2}.

We refer back to figures \ref{fig:image_base} and \ref{fig:epochs} to answer \textbf{Q3}. The performance of Siamese network with no STN is certainly better than with STN especially in lesser number of epochs although the difference in performance begins to shrink with the increase in the number of epochs. This is expected since STN, being a separate convolutional network in itself, takes longer number of epochs to train. Due to this steady increase in performance of Siamese network with STN we can answer \textbf{Q3} affirmatively. Our architecture can effectively handle the problem of rotational variance.

\begin{figure}[h!]
    \centering
    \includegraphics[width=\columnwidth]{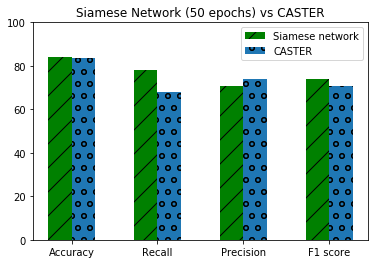}
    \caption{Comparison of our method (using images) with CASTER (using SMILES strings).}
    \label{fig:siamvscas}
\end{figure}

Figure \ref{fig:siamvscas} shows the result of our method when compared to a recent state-of-the-art method, CASTER. Our method outperforms CASTER across majority of metrics thereby proving the effectiveness of our approach in identifying drug-drug interactions. We show that using molecular structure images directly in a deep learning framework can result in a better/on-par performance than using \emph{lossy} string based representations. This answers \textbf{Q4}.

\begin{figure}[h!]
    \centering
    \includegraphics[width=\columnwidth]{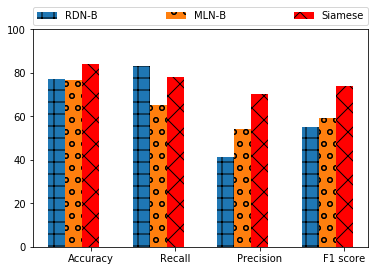}
    \caption{Comparison of our method (using images) with RDN-Boost and MLN-Boost (using relational data).}
    \label{fig:rel_base}
\end{figure}

Figure \ref{fig:rel_base} shows the result of comparing our method (Siamese network \textbf{without} STN trained for 50 epochs) to the state-of-the-art statistical relational learning baselines. Our method outperforms both the boosted methods across majority of the metrics. Note that similar to the results obtained when comparing with image based methods (figure \ref{fig:image_base}), although the recall of MLN-Boost is higher than the Siamese network, an accompanying low precision score shows a higher rate of false positives. This shows that using the molecular structural images directly can result in a better link prediction performance than using the data for the same drugs in a relational setting. This answers \textbf{Q5}.

An ideal predictor can use all the heterogeneous data types of the drugs considered i.e. images, string based representation and relational representation. We propose an initial sketch of such a model and leave it as future work. A graph convolutional network (GCN) \cite{kipf2016semi} is a type of graph neural network that extends the neural network models to be principally applied on graph data sets. A GCN makes use of a node feature matrix and the graph adjacency matrix to propagate functional values akin to a neural network to accomplish link prediction and node classification tasks. We propose a \emph{heterogeneous GCN} where heterogeneous data types available to us can be used to obtain the feature and adjacency matrix to be fed to the GCN. For example, relational data can be used to learn lifted rules which can then be grounded and the counts of the satisfied groundings can form a more richer and informed feature matrix than simple node features. A combination of the distances between the images and the string representation can form a more informed adjacency matrix and we can solve the drug-drug interaction problem as a link prediction problem. 

All the above reported results use euclidean distance as the metric to be used in contrastive loss while training the Siamese network ($D_P$ in equation \ref{eq}). To answer \textbf{Q6} we use 3 more distance metrics to be used inside the contrastive loss. These metrics are:
\begin{enumerate}
    \item \textbf{Manhattan distance:} This is the 1 norm distance between vectors i.e. the sum of absolute difference of the components of the vectors and is defined as $D_P={|F_P(X_1) - F_P(X_2)|_1}$.
    \item \textbf{Hellinger distance:} is a close relative of euclidean distance and is used to find the distance between 2 probability distributions. The Hellinger distance is given as \\$D_P=\sqrt{2 \sum (\sqrt{\frac{X_1}{\bar{X_1}}}-\sqrt{\frac{X_2}{\bar{X_2}}})^2}$.
    \item \textbf{Jaccard distance:} can be calculated in between binary segmentation of the input images and is given as $D_P={\frac{|X_1 \cap X_2|}{|X_1 \cup X_2|}}$.
\end{enumerate}

\begin{table*}[!ht]
    \centering
    \caption{Effect of choice of distance metric on the prediction performance.}
    \label{tab:results_dist}
    \begin{tabular}{|c|c|c|c|c|c|}
        \hline
         Distance Metric & Number of Epochs & Accuracy & Recall & Precision & F1\\
        \hline
        \multirow{3}{*}{Manhattan distance} & 20 & 0.817 & 0.849 & 0.645 & 0.733\\
        & 30 & 0.828 & 0.866 & 0.665 & \textbf{0.752}\\
        & 50 & 0.806 & 0.828 & 0.634 & 0.718\\
        \hline
        \multirow{3}{*}{Hellinger distance} & 20 & 0.300 & \textbf{1.0} & 0.297 & 0.461\\
        & 30 & 0.297 & \textbf{1.0} & 0.297 & 0.458\\
        & 50 & 0.295 & \textbf{1.0} & 0.295 & 0.456\\
        \hline
        \multirow{3}{*}{Jaccard distance} & 20 & 0.703 & 0.01 & 0.427 & 0.02\\
        & 30 & 0.703 & 0.0 & 0.7 & 0.0\\
        & 50 & 0.703 & 0.0 & \textbf{1.0} & 0.0\\
        \hline
        \multirow{3}{*}{Euclidean distance} & 20 & 0.822 & 0.835 & 0.657 & 0.735\\
        & 30 & 0.832 & 0.849 & 0.669 & 0.748\\
        & 50 & \textbf{0.839} & 0.78 & 0.705 & 0.741\\
        \hline
    \end{tabular}
\end{table*}

\begin{table*}[!ht]
    \centering
    \caption{Effect of choice of optimization function on the prediction performance.}
    \label{tab:results_opt}
    \begin{tabular}{|c|c|c|c|c|c|}
        \hline
         Optimization function & Number of Epochs & Accuracy & Recall & Precision & F1 Score\\
        \hline
        \multirow{3}{*}{RMSprop \citep{hinton2012lecture}} & 20 & 0.822 & 0.672 & 0.715 & 0.693\\
        & 30 & 0.770 & \textbf{0.877} & 0.576 & 0.696\\
        & 50 & 0.816 & 0.640 & 0.707 & 0.672\\
        \hline
        \multirow{3}{*}{Adadelta \citep{zeiler2012adadelta}} & 20 & 0.721 & 0.138 & 0.667 & 0.229\\
        & 30 & 0.827 & 0.813 & 0.672 & 0.735\\
        & 50 & \textbf{0.851} & 0.831 & 0.707 & \textbf{0.764}\\
        \hline
        \multirow{3}{*}{Nadam \citep{dozat2016incorporating}} & 20 & 0.812 & 0.852 & 0.639 & 0.730\\
        & 30 & 0.828 & 0.833 & 0.668 & 0.742\\
        & 50 & 0.848 & 0.790 & \textbf{0.721} & 0.754\\
        \hline
        \multirow{3}{*}{Adam \citep{kingma2014adam}} & 20 & 0.822 & 0.835 & 0.657 & 0.735\\
        & 30 & 0.832 & 0.849 & 0.669 & 0.748\\
        & 50 & 0.839 & 0.780 & 0.705 & 0.741\\
        \hline
    \end{tabular}
\end{table*}

Table \ref{tab:results_dist} shows the effect of using different distance metrics within the contrastive loss on the performance of the Siamese architecture (without STN). The results show that the use of euclidean and Manhattan distance as the metric in the contrastive loss perform similarly and outperform Hellinger and Jaccard distance by huge margins. Although the recall values using Hellinger distance and precision values using Jaccard distance, at 50 epochs, are perfect i.e. equal to 1, the respective precision and recall values in both the distances are very low thereby showing that using these distances in the contrastive loss leads to poor performance. This answers \textbf{Q6}.

Table \ref{tab:results_opt} shows the effect of using different optimization functions (RMSProp, Adadelta and Nadam) to optimize the Siamese network with increasing number of epochs. The last row in table \ref{tab:results_dist} shows the results with using Adam as the optimization function with increasing number of epochs. We include that row in table \ref{tab:results_opt} for more clarity.

The results vary widely with respect to the optimization function used with an increase in the performance wrt the increasing number of epochs in case of Adadelta \citep{zeiler2012adadelta} optimization function. In case of the other 3 optimization functions, interestingly, we note that there is a drop in recall when we go from 30 to 50 epochs. This shows that the choice of the optimization function does play a big part in the prediction performance thereby answering \textbf{Q7}.

\section{Conclusion}
In this work we focus on using the molecular images of the drugs in a pairwise fashion and feeding them to a rotation-invariant Siamese architecture to predict whether two drugs interact with each other. Our evaluations on the drug images obtained from PubChem database establish the superiority of our proposed approach, which is distinct from current approaches that generally uses SMILES and  \textbf{SM}iles \textbf{AR}bitrary \textbf{T}arget \textbf{S}pecification (\textbf{SMARTS}) strings \citep{sayle19971st}.

Combining our previous work \citep{dhami2018drug} that used different similarity measures obtained from a directed graph of known chemical reactions between drugs and enzymes, transporters and inhibitors as well as the structure of the drugs in the form of SMILES and SMARTS strings and the current work which uses images of the drug structure is a natural next step. Also refining the Siamese architecture and feeding more drug images to the network are an interesting area of future work.

\bibliographystyle{ACM-Reference-Format}
\bibliography{test}

\end{document}